%% file: acl_latex.tex
\pdfoutput=1

\documentclass[11pt]{article}

\usepackage[final]{acl}

\usepackage{times}
\usepackage{latexsym}

\input{preamble}

\input{sec/math_commands}

\input{sec/notations}

\usepackage[T1]{fontenc}

\usepackage[utf8]{inputenc}

\usepackage{microtype}

\usepackage{inconsolata}

\usepackage{graphicx}

\usepackage{hyperref}
\renewcommand{\sectionautorefname}{\S}
\usepackage{amssymb}

\title{Can Hallucination Correction Improve Video-Language Alignment?}

\author{
  Lingjun Zhao$^{1}$\thanks{This work was partially conducted during an internship at Honda Research Institute.}, 
  Mingyang Xie$^{1}$, \\
  \textbf{Paola Cascante-Bonilla}$^{1,2}$,  
  \textbf{Hal Daum\'e III}$^{1}$, 
  \textbf{Kwonjoon Lee}$^{3}$ \\ 
  $^{1}$University of Maryland, College Park \
  $^{2}$Stony Brook University \
  $^{3}$Honda Research Institute \\ 
  \texttt{lzhao123@umd.edu} 
}

\begin{document}
\maketitle

\input{sec/0_abstract}



\input{sec/1_intro}

\input{sec/2_related_work_main}

\input{sec/3_method}

\input{sec/4_results}
\input{sec/5_analysis}
\input{sec/6_conclusion}

\section*{Limitations}
\input{sec/limitations}

\section*{Acknowledgements}
\input{sec/acknowledgement}

\bibliography{custom}

\clearpage
\onecolumn

\appendix
\input{sec/Appendix}
\input{sec/2_related_work2}

\end{document}

%% file: preamble.tex
\usepackage{multirow}
\usepackage{xcolor}
\usepackage{colortbl}
\usepackage{adjustbox}
\usepackage{hhline}
\usepackage{tabu}
\usepackage{makecell}
\usepackage{subcaption}
\usepackage{bm}
\usepackage{arydshln, collcell}

\usepackage{booktabs}
\usepackage{color}
\usepackage{subcaption}
\usepackage{pifont}
\usepackage{paralist}
\usepackage{threeparttable}
\usepackage{scalerel,xparse}

\usepackage{enumitem}

\definecolor{Gray}{gray}{0.5}
\definecolor{LGray}{gray}{0.9}
\definecolor{darkblue}{RGB}{94,110,186}
\definecolor{darkGreen}{RGB}{92, 148, 110}
\definecolor{myblue}{RGB}{14, 121, 178}
\definecolor{myred}{RGB}{192, 0, 0}



%% file: sec/math_commands.tex
\usepackage{amsmath,amsfonts,bm}

\def\eqref#1{equation~\ref{#1}}

\def\1{\bm{1}}

\DeclareMathAlphabet{\mathsfit}{\encodingdefault}{\sfdefault}{m}{sl}
\SetMathAlphabet{\mathsfit}{bold}{\encodingdefault}{\sfdefault}{bx}{n}

%% file: sec/notations.tex
\definecolor{lightgray}{HTML}{808080}
\definecolor{perfup}{HTML}{008000}
\definecolor{perfdown}{HTML}{1d7b21}
\definecolor{myred}{HTML}{e60035}

\newcommand\redsout{\bgroup\markoverwith{\textcolor{red}{\rule[0.5ex]{2pt}{2pt}}}\ULon}
\newcommand\bluesin{\bgroup\markoverwith{\textcolor{green}{\rule[-0.5ex]{2pt}{2pt}}}\ULon}
\newcommand\redsin{\bgroup\markoverwith{\textcolor{red}{\rule[-0.5ex]{2pt}{2pt}}}\ULon}

\renewcommand{\sectionautorefname}{\S\kern-0.2em}
\renewcommand{\subsectionautorefname}{\S\kern-0.2em}

\definecolor{lightgreen}{HTML}{1d7b21}

\newcommand{\textpr}[1]{\textcolor{violet}{\texttt{#1}}}

\renewcommand{\vec}[1]{\boldsymbol{#1}}
\newcommand{\video}{\vec V}
\newcommand{\instr}{\vec Q}
\newcommand{\answer}{\vec A}
\newcommand{\des}{\vec{\hat{W}}}
\newcommand{\sent}{\vec W}
\newcommand{\mask}{\bar{\vec{W}} }

\newcommand{\method}{\text{HACA}}

%% file: sec/0_abstract.tex
\begin{abstract}
Large Vision-Language Models often generate hallucinated content that is not grounded in its visual inputs.
While prior work focuses on mitigating hallucinations, we instead explore leveraging hallucination correction as a training objective to improve video-language alignment.
We introduce \method{}, a self-training framework learning to correct hallucinations in descriptions that do not align with the video content.
By identifying and correcting inconsistencies, \method{} enhances the model’s ability to align video and textual representations for spatio-temporal reasoning.
Our experimental results show consistent gains in video-caption binding 
and text-to-video retrieval %
tasks, demonstrating that hallucination correction-inspired tasks serve as an effective strategy for improving vision and language alignment.

\end{abstract}

%% file: sec/1_intro.tex
\section{Introduction}
\label{sec:intro}

Aligning representations across modalities involves creating joint embeddings that map visual and linguistic features to a shared space, enabling the model to assess their similarity.
This is crucial for tasks including cross-modal retrieval \cite{xu2016msr}, mitigating hallucinations \cite{jiang2024hallucination},
compositional reasoning \cite{cascante2024natural}, 
visual-to-text generation \cite{blip2}, visual question answering \cite{shen2022how}, and visual-language navigation \cite{zhao2021evaluation}.
While progress has been made in aligning image representations with text \cite{li2020oscar, zeng2021multi, wang2023image, lin2025evaluating}, advancements in video-language alignment remain limited.
Videos pose unique challenges due to their rich spatio-temporal information, involving multiple entities and scenes that dynamically interact and change over time.
Video-language models (Video-LLMs in \autoref{sec:prelim}) can compute alignment scores \cite{lin2023video, li2023videochat} but struggle to distinguish between similar videos and descriptions \cite{park2022exposing, wang2023paxion, saravanan2024velociti},
as illustrated in \autoref{fig:haca}. 
One promising approach is to fine-tune Video-LLMs on entailment tasks using similar captions~\cite{bansal2024videocon}, where the model is prompted to answer \textpr{Yes} or \textpr{No} to whether a video is aligned with a given caption (\autoref{sec:prelim}). 
However, using a single binary label as a learning signal fails to indicate which parts of the description misalign with the video. 
\citet{bansal2024videocon} generates natural language explanations for mismatches but requires costly dataset construction with additional models and annotations.

\input{figures/haca}

To this end, we introduce \textbf{\method{}}, a self-training framework grounded in \textbf{HA}llucination \textbf{C}orrection for video-language \textbf{A}lignment (\autoref{sec:hallucination_correction}).
\textit{Hallucination} (or confabulation) refers to a mismatch between textual descriptions and the corresponding factual content of an image or video \cite{liu2024survey}.
\method{} requires the model to predict whether a description entails the video content. If the description does not align, the model corrects the hallucinations to better match the video.
Instead of relying solely on a binary entailment label, \method{} uses hallucination correction as a finer-grained learning signal to enhance the alignment of video and language representations.
Given that misalignment between modalities is a key factor in hallucination \cite{biten2022let, sun-etal-2024-aligning}, we hypothesize that introducing a hallucination correction task can improve video-language alignment.
\method{} also requires no external models or annotations beyond the ground-truth video description. 
To further enhance \method{}, we introduce a masking correction task as data augmentation
(\autoref{sec:masking_correction}).

We fine-tune two Video-LLMs with \method{},
and evaluate these fine-tuned models in a zero-shot manner on two spatio-temporally challenging downstream tasks (\autoref{sec:exp_setup}): VELOCITI~\cite{saravanan2024velociti}, a video-caption binding dataset, and SSv2-Temporal \cite{sevilla2021only} and SSv2-Events \cite{bagad2023test}, which are text-to-video retrieval datasets emphasizing action recognition.
The models fine-tuned with \method{} outperform baseline models by up to 17.9\% accuracy and 5.7 mAP points, 
demonstrating that \method{} effectively improves video-text alignments, and generalizes beyond in-domain data (\autoref{sec:analysis}). 
Our code and data will be available upon acceptance.

%% file: figures/haca.tex
\begin{figure}[t]
    \centering
    \includegraphics[width=.5\textwidth]{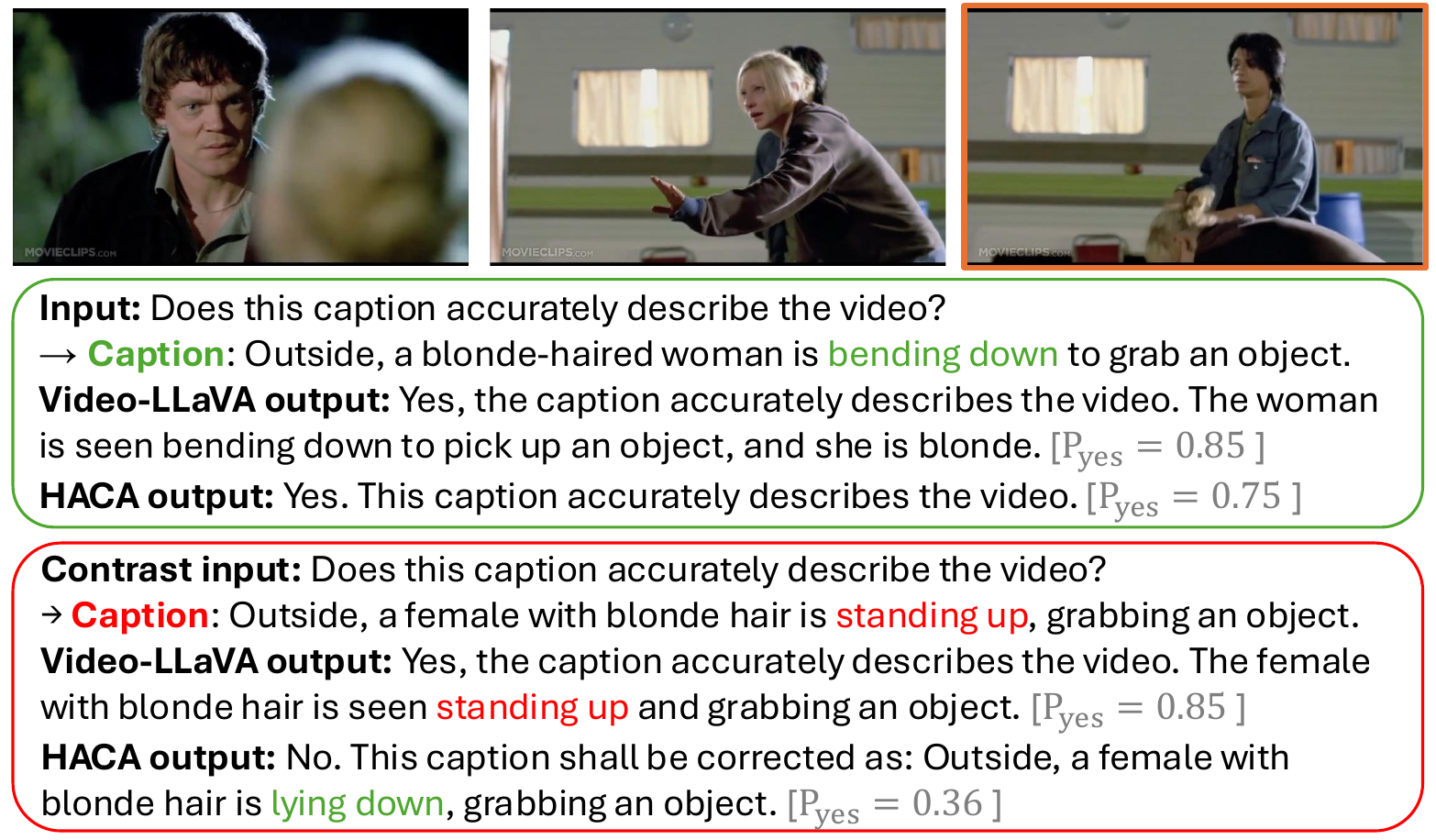} 
    \vspace{-0.5cm}
    \caption{
    Models tasked with determining whether a given video entails a caption, where the contrast caption closely resembles the correct one. 
    \method{} effectively differentiates between the correct caption (top) and the incorrect one (bottom), and corrects hallucination in the latter.
    In contrast, Video-LLaVA fails to distinguish between those captions or correct the hallucination.}
    \label{fig:haca}
    \vspace{-0.3cm}
\end{figure}

%% file: sec/2_related_work_main.tex
\section{Related Work}
\looseness=-1
In addition to the video-language alignment approaches discussed in \autoref{sec:intro}, several methods leverage a contrastive learning objective to learn a shared video-language embedding space \cite{xue2023clipvip, rasheed2023fine, girdhar2023imagebind, zhu2024languagebind}, and \citet{bagad2023test} further introduces a contrastive loss in Video-LLMs to enforce time-order consistency. 
However, most of these models lack robustness to semantically plausible manipulations \cite{park2022exposing}. 
\citet{yuksekgonul2022and} also finds that applying a contrastive objective to video-caption datasets does not promote the model’s ability to capture fine-grained details. 
In contrast, our approach encourages Video-LLMs to capture more nuanced semantic mismatches by learning to correct hallucinations, extending beyond sentence-level hallucination detection.
More discussion is provided in \autoref{app:related_work}.

%% file: sec/3_method.tex
\input{figures/all_tasks}

\section{HACA: Hallucination Correction for Video-language Alignment}
To investigate whether hallucination correction can improve video-language alignment, we introduce \method{}, 
a fine-tuning objective for Video-LLM
as a sequence-to-sequence generation task.

\subsection{Preliminaries: Video-LLMs}
\label{sec:prelim}

Video-LLMs typically consist of three parts: i) a visual encoder to map images and videos to visual representations; ii)
an LLM that takes text instructions as inputs to generate text responses; and iii) an adapter between visual and text representations. Our approach finetunes (ii) the text decoder and (iii) the adapter, freezing (i) the visual encoder.

\paragraph{Pre-training.}
A Video-LLM $M_\theta$ parameterized by $\theta$ takes a textual question or instruction $\instr$, a video $\video$ as input, and generates a text response $\answer$ = $(\answer_1, \answer_2, ..., \answer_T)$ autoregressively using a decoder-based language model (LLM) as output, by estimating a conditional distribution  $M(\answer \mid \instr, \video)$.
This is achieved by training the model using the maximum-likelihood estimation (MLE) objective:
\begin{equation}
 L(\theta) = \sum_{\mathcal{D}_{\textrm{train}}}  \sum_{t=1}^{T} \log M_{\theta} (\answer_t \mid \answer_{<t}, \instr, \video)
 \label{equation:videollm_training}
\end{equation}
where $\answer_t$ is $t$-th word of the text response, and $\answer_{<t}$ are the first $t - 1$ words of the response.
The dataset $\mathcal{D}_{\textrm{train}}$ consists of samples in the form $(\instr, \answer, \video)$.

\paragraph{Fine-tuning with entailment.}
\label{method:prior_finetuning}
Following \citet{bansal2024videocon}, we finetune the Video-LLM using an entailment task, where
the text input $\instr$ is formatted as an entailment question %
as $\instr(\sent) =$ \textpr{Does this caption accurately describe the video? Caption: \{$\sent$\}}.
In this task, the output of the model $\answer$ is \textpr{Yes} or \textpr{No} (\autoref{fig:tasks} (a)).
Given a dataset $D_{train}$ consisting of ground-truth answers $\answer$ for $\instr(\sent)$ and $\video$, the model is fine-tuned to have a better estimation of $M_\theta(\textpr{Yes} ~|~ \instr(\sent), \video) $ and $M_{\theta}(\textpr{No} ~|~ \instr(\sent), \video)$ using the MLE objective:
\begin{equation}
 L_{ent}(\theta) = \sum_ {\mathcal{D}_{\textrm{train}}}   \log M_{\theta} (\answer \mid \instr(\sent), \video)
 \label{equation:entail_training}
\end{equation}

\subsection{Learning from Hallucination Correction}
\label{sec:hallucination_correction}

Building on the work of \citet{bansal2024videocon}, the Video-LLM takes the question $\instr$ and the video $\video$ as input to determine whether a text description $\sent$ entails the video (similar to~\autoref{method:prior_finetuning}). 
However, in our setting, if $\sent$ does not entail $\video$, the model generates a \textit{corrected} caption $\des=(w_1, w_2, ..., w_n)$ to align the description with the video content.
During fine-tuning, if $\sent$ entails $\video$, the model is trained to generate the response as
$\answer(\textpr{Yes}) =$ \textpr{Yes, the caption accurately describes the video}.
If $\sent$ does not entail $\video$, the model is trained to generate a corrected description $\des$ as its response, formatted as
$\answer(\textpr{No}, \des) =$ \textpr{No. This caption shall be corrected as: \{$\des$\}}.
We show an example in \autoref{fig:tasks} (b). In contrast to finetuning using \textit{entailment} only, 
our \textit{hallucination correction} objective trains the model 
to have better estimation of $M_\theta(\answer(\textpr{Yes}) ~|~ \instr(\sent), \video) $ and $M_\theta(\answer(\textpr{No}, \des) | \instr(\sent), \video) $.

Instead of using ~\autoref{equation:entail_training}, given a training dataset $\mathcal{D}_{\textrm{train}}$ that consists of video $\video$ and ground-truth text description $\des$ pairs,
we fine-tune the Video-LLM using the MLE objective:
\begin{equation}
 L_{c}(\theta) = \sum_{\mathcal{D}_{\textrm{train}}} \sum_{t=1}^{T} \log M_{\theta} (\answer_t \mid \answer_{<t}, \instr(\sent), \video)
 \label{equation:haca}
\end{equation}
where $\answer_t$ is the $t$-th word of the text response of $\answer(\textpr{Yes})$ or $\answer$(\textpr{No}, $\des$), and $\answer_{<t}$ is the first $t - 1$ words of the text response.

\subsection{Masking Correction as Augmentation}
\label{sec:masking_correction}

We also incorporate a masking correction task as data augmentation (\autoref{fig:tasks} (c)),
where an instruction $\instr$ prompts the Video-LLM to make corrections to a masked caption $\mask$, teaching the model to generate a corrected caption that contains a sequence of words $\des=(w_1, w_2, ..., w_n)$ as its answer
by estimating conditional probability $M(\des \mid \instr(\mask), \video)$.
Specifically, $\instr$ is a function that formats the text instruction as 
$\instr(\mask) = $ \textpr{Please correct this caption to accurately describe the video. Caption: $\mask$},
where $\mask$ is masked from $\des$: $\mask = (w_1, \textpr{[MASK]}, ..., w_n )$, by randomly masking 45\% of the content words in the ground truth video description $\des$. 

We finetune the model using two objectives: the MLE objective to estimate the probability for masking correction $M_\theta(\des| \instr(\mask), \video)$, and the \method{} objective (\autoref{equation:haca}).
The model is tasked with providing responses corresponding to different instructions.

%% file: figures/all_tasks.tex
\begin{figure*}[t]
    \centering
    \includegraphics[width=.99\textwidth]{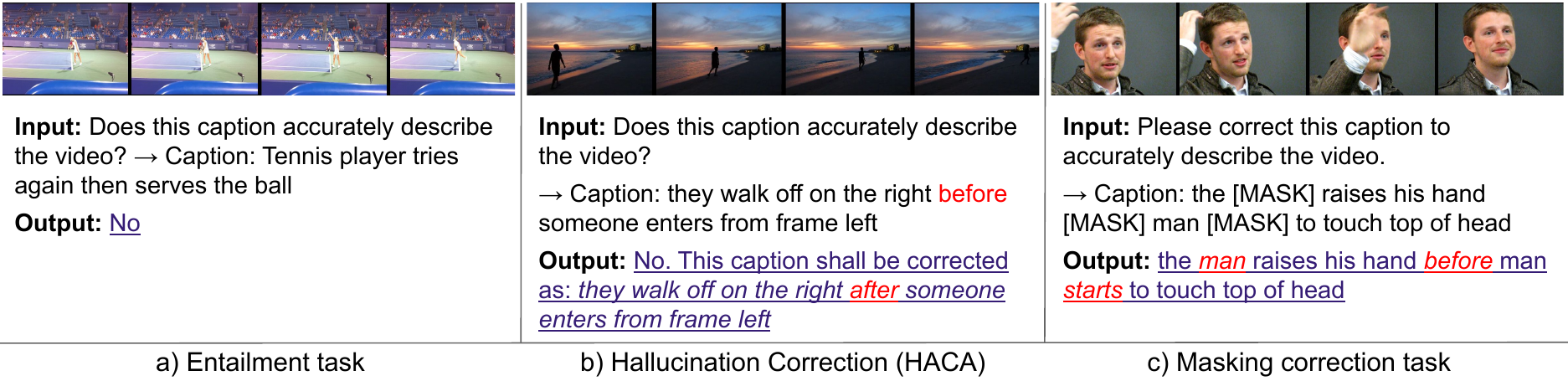}
    \vspace{-0.2cm}
    \caption{Example of different finetuning objectives. 
    The first column shows an example of the baseline \textit{entailment} task.
    The second column shows an example of our proposed \textit{\method{}} task, where we finetune the model to output hallucination correction to justify the response. 
    The third column shows an example of the \textit{masking correction} task, where we input a masked version of the video description and finetune the model to predict the corrected one. 
    }
    \vspace{-0.3cm}
    \label{fig:tasks}
\end{figure*}

%% file: sec/4_results.tex
\section{Experimental Setup}
\label{sec:exp_setup}

\looseness=-1
\paragraph{Data (detailed in \autoref{app:datasets}).}

We train \method{} using videos and their ground-truth and contrastive descriptions from VideoCon \cite{bansal2024videocon}, generating 115,536 (video, description, correction) triplets for training and 8,312 for validation, which is used for model selection. Synthetic contrast captions are also used to fine-tune the \textit{baseline} entailment task with the same dataset sizes.

We evaluate our trained models on text-to-video retrieval using the temporally-challenging SSv2-Temporal \cite{sevilla2021only} and action-intensive SSv2-Events \cite{bagad2023test}  datasets.
Additionally, we evaluate our models on 
compositional ability over time using the VELOCITI benchmark \cite{saravanan2024velociti}. Each video in the dataset includes a correct caption and an incorrect one.

\paragraph{Baselines.}
(i) \textbf{Pretrained Video-LLMs}:
we employ two pre-trained models with different architectures, \textit{Video-LLaVA}~\cite{lin2023video} 
and
\textit{VideoChat2}~\cite{li2023videochat}. 
More details in \autoref{app:pretrained_vlm}.
(ii) \textbf{Entailment}:
we fine-tune the pretrained Video-LLMs using the entailment task described in \autoref{sec:prelim}. 
More details about the implementation are in \autoref{app:implementation}.

\paragraph{Evaluation metrics.}

We report the accuracy on the VELOCITI benchmark as the proportion of examples in which the positive video-caption pair receives a higher \textpr{Yes} entailment probability than the corresponding negative video-caption pair.
For SSv2, we compute \textpr{Yes} probabilities for each text-video pair, rank their scores, and report mean Average Precision (mAP).

\input{figures/ssv2_chart}

\input{tables/velociti}

%% file: figures/ssv2_chart.tex
\begin{figure}[t!]
\centering
\includegraphics[width=0.45\textwidth]{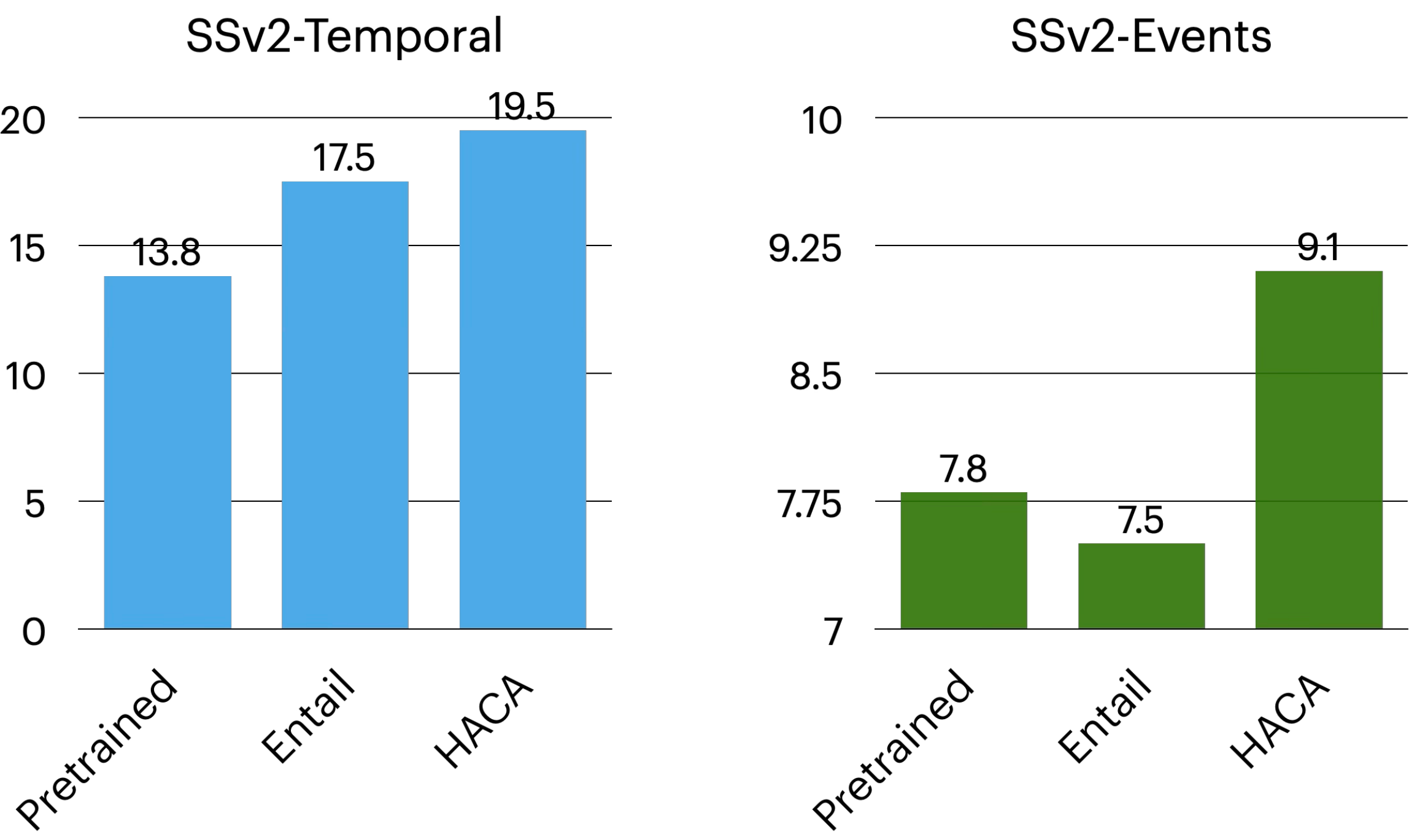}
\vspace{-0.3cm}
\caption{Mean Average Precision (mAP) scores for pretrained Video-LLaVA and models fine-tuned using various methods on zero-shot text-to-video retrieval tasks. }
\label{fig:ssv2}
\vspace{-0.2cm}
\end{figure}

%% file: tables/velociti.tex
\begin{table}[t]
\centering
\begin{adjustbox}{width=\columnwidth}
\begin{tabular}{@{}l@{~~}c@{ }c@{ }c@{~~}c@{ }c@{ }c@{~~}c@{~~}>{\columncolor{gray!10}}c@{}}
\toprule
 & \multicolumn{3}{c}{\textbf{Agent Tests}}               & \multicolumn{3}{c}{\textbf{Action Tests}} & & \\ 
\textbf{Model} & Iden~~          & Bind~          & Coref         & Adv           & Bind~          & Modif         &  \textbf{Chrono}                       &    \textbf{Avg}                  \\
\midrule
Random          & 50.0          & 50.0          & 50.0          & 50.0          & 50.0          & 50.0          & 50.0 & 50.0          \\
Human           & 94.7          & 93.3          & 96.0          & 100.0         & 92.7          & 91.3          & 93.3 & 94.4          \\
\midrule
\textcolor{gray}{CLIP-ViP {\scriptsize B/32}} & \textcolor{gray}{75.3} & \textcolor{gray}{52.4} & \textcolor{gray}{55.7} & \textcolor{gray}{70.2} & \textcolor{gray}{53.5} & \textcolor{gray}{51.2} & \textcolor{gray}{48.5} & \textcolor{gray}{58.1} \\
\textcolor{gray}{ViFi-CLIP {\scriptsize B/16}} & \textcolor{gray}{82.3} & \textcolor{gray}{58.7} & \textcolor{gray}{54.6} & \textcolor{gray}{63.0} & \textcolor{gray}{59.3} & \textcolor{gray}{60.5} & \textcolor{gray}{49.8} & \textcolor{gray}{61.2} \\
\midrule
mPLUG-V & 43.0 & 31.9 & 51.7 & 65.0 & 42.0 & 49.6 & 41.3 & 46.3 \\
PLLaVA & 68.6 & 43.3 & {60.5} & 62.4 & 46.6 & 56.0 & 49.6 & 55.3 \\
VideoCon & 67.4 & 44.6 & 50.0 & 73.0 & 51.1 & 63.2 & 45.6 & 56.4 \\

\arrayrulecolor{gray!50}
\hdashline
\rowcolor[HTML]{F2F3F4} 
Video-LLaVA & 74.1 & 50.4 & \textbf{60.1} & 63.6 & 47.0 & 47.9 & 56.0 & 57.0  \\
\hdashline
\arrayrulecolor{black}
$+$ Entail & 73.7  & 59.7  & 55.5  & 68.4  & 57.3  & 64.0  & 57.3  & 62.3 \\
$+$ \method{} & 80.3 & \textbf{62.6} & 57.9 & \textbf{72.6} & \textbf{60.0} & \textbf{65.8} & 54.5 & 64.8 \\
$+$ \method{}+Mask & \textbf{82.7} & 62.1 & 57.9 & 71.8 & 59.0 & 64.8 & \textbf{57.9} & \textbf{65.2} \\
\arrayrulecolor{gray!50}
\hdashline
\rowcolor[HTML]{F2F3F4} 
VideoChat2         & 76.8 & 54.4 & 53.1 & 56.0 & 46.2 & 59.3 & \textbf{54.7} & 57.2 \\
\hdashline
\arrayrulecolor{black}
$+$ Entail         & 59.7 & 56.9 & 55.5 & 62.2 & 53.0 & 50.1 & 53.9 & 55.9\\
$+$ \method{}      & 77.2 & \textbf{60.4} & 56.4 & 65.8 & \textbf{55.0} & 61.7 & 53.7 & 61.5\\
$+$ \method{}+Mask & \textbf{79.1} & 59.7 & \textbf{56.9} & \textbf{68.2} & 54.6 & \textbf{66.9} & 51.1 & \textbf{62.4} \\
\bottomrule
\end{tabular}
\end{adjustbox}
\caption{
Zero-shot accuracy on VELOCITI for models trained with the baseline entailment task, our proposed \method~objective, and other contrastive (CLIP-ViP~\cite{xue2023clipvip}, ViFi-CLIP~\cite{rasheed2023fine}) and generative (mPLUG-V~\cite{ye2023mplug}, PLLaVA~\cite{Xu2024PLLaVAP}) models.
}
\label{table:velociti}
\vspace{-0.3cm}
\end{table}

%% file: sec/5_analysis.tex
\section{Analysis}
\label{sec:analysis}
\looseness=-1

\paragraph{Performance on text-to-video retrieval.} 
\method{} consistently outperforms both the pretrained model and the entailment fine-tuned model, as illustrated in \autoref{fig:ssv2}. This demonstrates \method{}'s ability to effectively capture the rich temporal information present in videos. On the SSv2-Events dataset, while the entailment objective yields performance comparable to the pretrained Video-LLaVA, \method{} achieves better results on this action-intensive dataset, despite being fine-tuned on the same amount of data.
Additional comparisons with other models are provided in \autoref{app:ablation}.

\paragraph{Performance on video-language binding.} 
\autoref{table:velociti} shows that, on average, both Video-LLaVA and VideoChat2 fine-tuned with the \method{} objective outperform the pre-trained models and those fine-tuned with the entailment objective. 
Masking correction further boosts performance through data augmentation.
The \textit{Agent Coref} test evaluates a model's ability to link events to specific agents, a misalignment type absent in the VideoCon dataset, where actions are always tied to one agent. Consequently, the pretrained Video-LLaVA outperforms its fine-tuned versions, with \method{} marginally exceeding the entailment baseline.
The \textit{Chrono} test measures a model's ability to detect reversed event order. While VideoCon includes such data, our results show that models fine-tuned on the entailment objective perform similarly to the pretrained model. Although \method{} slightly underperforms the entailment objective, it excels on SSv2-Events, involving multiple events.

\method{} consistently outperforms baseline models in all \textit{Action} tests: \textit{Action Adv} (replacing an action with one not in the video), \textit{Action Bind} (replacing an action within the same video), and \textit{Action Modif} (replacing the manner with a plausible modifier). 
This highlights \method{}'s robust ability to distinguish actions in videos, requiring understanding of complex spatio-temporal relationships between the video and its description.
\method{} also excels in \textit{Agent Iden}
and \textit{Agent Bind},
showcasing its effectiveness in identifying and binding entities through the right relationship.

\paragraph{Qualitative examples.}
\input{figures/qualitative_main}
\autoref{fig:qualtitative_main} presents an example where \method{} outperforms the entailment baseline on the VELOCITI dataset, and delivers accurate corrections. Additional qualitative examples are provided in \autoref{app:qualitative_analysis}.

\paragraph{\method{} does not hinder question answering.}
To assess whether fine-tuning with the \method{} objective affects the multi-task capabilities of Video-LLMs, we conduct a zero-shot evaluation on the MSRVTT-QA dataset \cite{xu2016msr} using GPT-3.5-turbo. The results, presented in \autoref{tab:msrvtt_vqa}, are based on a subset of 7,000 samples (10\% of the dataset) due to budget constraints. The GPT-evaluated score for the pretrained Video-LLaVA model aligns with previously reported values \cite{lin2023video}.
As shown in the table, the \method{}-finetuned model performs comparably to the pretrained Video-LLaVA model, despite not involving explicit QA-specific finetuning. In contrast, fine-tuning with the entailment objective alone leads to a significant performance drop on MSRVTT-QA. A possible explanation is that optimizing for a single entailment label may impair the language generation capabilities of video-language models.

\begin{table}[h]
    \centering
    \small
    \begin{tabular}{lc}
        \toprule
        \textbf{Model}       & \textbf{GPT Score} \\
        \midrule
        Video-LLaVA & 3.5 \\
        + Entail     & 2.8 \\
        + HACA       & 3.4 \\
        \bottomrule
    \end{tabular}
    \caption{Zero-shot GPT-assessed score on MSRVTT-QA for the model trained with the baseline entailment task, and our proposed \method{} objective. GPT-assessed scores ranges from 0 to 5.}
    \label{tab:msrvtt_vqa}
\end{table}

%% file: figures/qualitative_main.tex
\begin{figure}[t!]
\centering
\includegraphics[width=0.5\textwidth]{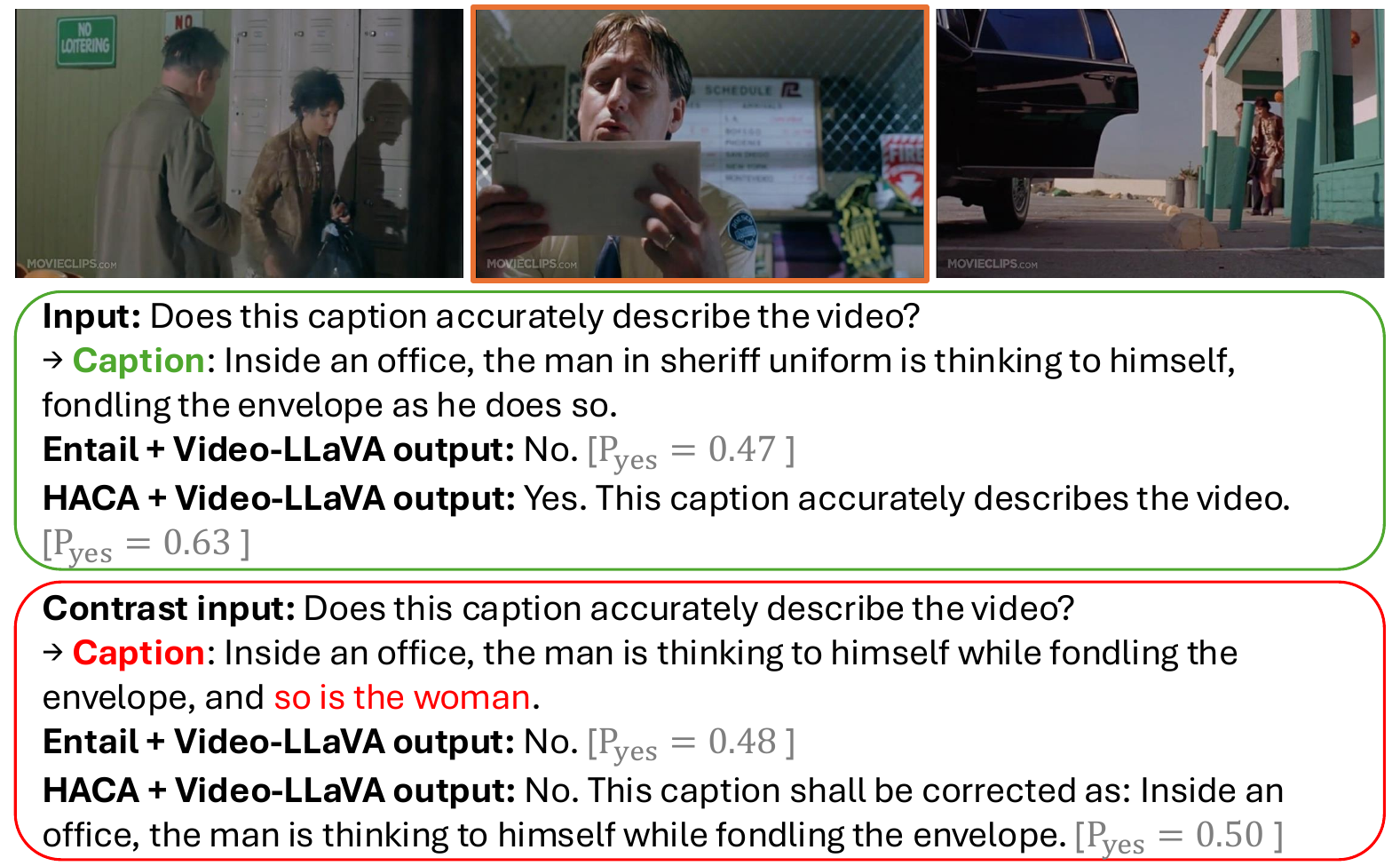}
\vspace{-0.7cm}
\caption{Success on binding and correction: \method{} effectively assigns higher entailment probability $P_{yes}$ to the correct caption (top) than the incorrect one (bottom), unlike the entailment-finetuned model. \method{} also accurately corrects the incorrect caption in its output.}
\vspace{-0.45cm}
\label{fig:qualtitative_main}
\end{figure}

%% file: sec/6_conclusion.tex
\section{Conclusion}

Video understanding through language is vital for applications like human-robot interaction and autonomous driving. We propose a novel approach to enhance video-language alignment by connecting it to the hallucination problem in visual-language models, paving the way for future advancements.

%% file: sec/limitations.tex
Our proposed method assumes the availability of ground-truth video caption annotations for fine-tuning using hallucination correction. 
Additionally, the method assumes a clear separation between the parameters of the video representations and those of the language model, as we freeze the video encoder parameters during fine-tuning to align video-language representations. 
Another limitation is that our approach has not been evaluated on long videos, due to the limitation of computational resources. We envision future work in this direction.

%% file: sec/acknowledgement.tex
This material is based upon work partially supported by the NSF under Grant No. 2229885 (NSF Institute for Trustworthy AI in Law and Society, TRAILS).
We thank Haoqiang Kang for guidance on optimizing the computation speed of fine-tuning video-language models.

%% file: sec/Appendix.tex
\section{Appendices}
\label{sec:appendix}

\subsection{Implementation Details}
\label{app:implementation}

\paragraph{Fine-tuning from pretrained models.}
We use the visual representations for video and language model embeddings pretrained from Video-LLMs to perform instruction fine-tuning using different objectives, including \method{} (\autoref{sec:hallucination_correction}), entailment (\autoref{sec:prelim}), and masking correction (\autoref{sec:masking_correction}).
During finetuning, visual representations are frozen, and the embeddings from the visual-text adapter layers and LLM are learnable.

\paragraph{Hyperparameters and computation.}

For Video-LLaVA, we finetune our models for $3$ epochs, using a learning rate of $2e^{-4}$ and AdamW optimizer. We also use a LoRA adapter~\cite{hu2022lora} of rank $128$ and alpha $256$. Since we freeze the video encoder, the number of trainable parameters is significantly reduced to $241$M for Video-LLaVA. The number of video frames processed per video is $8$, with a batch size of $8$, using $2$ RTXA6000 GPUs, for a total of $\sim 72$ hours.

For VideoChat2, we finetune our models for $3$ epochs, using a learning rate of $2e^{-5}$ and AdamW optimizer. We use a LoRA adapter~\cite{hu2022lora} of rank $16$ and alpha $32$. We also freeze the visual encoders and reduce the number of trainable parameters to $193$M.  The number of video frames processed per video is $8$, with a batch size of $2$, using $1$ RTXA6000 GPU, for around $\sim 72$ hours.

\paragraph{Tools.}
We implement our models with Pytorch 2.0.1, Huggingface Transformers 4.31.0, scikit-learn 1.2.2. We use SciPy 1.6.0 to find content words from ground truth video description by excluding words with part-of-speech tags: AUX, SYM, DET, PUNCT.

\subsection{Ablation studies}
\label{app:ablation}

\paragraph{Performance on text-to-video retrieval.}
In \autoref{table:ssv2}, \method{} (\autoref{sec:hallucination_correction}) demonstrates competitive performance on SSv2 downstream tasks, surpassing the pretrained model by up to 5.7 mAP points and outperforming the model fine-tuned with the entailment task by up to 2.0 mAP points.
Masking correction augmentation typically enhances video-language alignment when jointly trained with \method{} or the entailment task.

\input{tables/ssv2}

\paragraph{Effect of different mask ratios.}
\autoref{table:mask_ratio} shows the performance when jointly finetuning Video-LLaVA using \method{} and masking correction task (\autoref{sec:masking_correction}) with different masking ratio.
The results indicate that using masking ratio of 45\% achieves higher average accuracy.

\input{tables/different_mask_ratio}

\paragraph{Comparing \method{} and natural language explanations.}
To assess the effectiveness of \method{} as a finetuning task, we compare it against natural language explanations (NLE) generated by external natural language inference models \cite{bansal2024videocon}, used alongside the entailment task. We fine-tune Video-LLaVA with both the entailment and NLE training objectives and report the results in \autoref{table:nle}. Our findings show that \method{} outperforms Video-LLaVA trained with entailment and NLE objectives, even without our proposed masking objective.

\input{tables/nle}

\subsection{Additional Qualitative Analysis}
\label{app:qualitative_analysis}

\input{figures/qualitative_app}

\autoref{fig:qualitative_app} shows additional success and failure cases of \method{} and the other models we tested.

\subsection{Pretrained Video-LLMs.}
\label{app:pretrained_vlm}
We use two pre-trained Video-LLMs with different model architectures.
\paragraph{Video-LLaVA.} Video-LLaVA~\cite{lin2023video} consists of LanguageBind~\cite{zhu2024languagebind} encoders for the visual inputs, a large language model~\cite{vicuna}, visual projection layers and a word embedding layer. It is finetuned via visual instruction tuning with 665k image-text pairs from LLaVA 1.5~\cite{llava} and a 100k video-text instruction set from Video-ChatGPT~\cite{videochatgpt}.
We use this model under their Apache License 2.0.

\paragraph{VideoChat2.} VideoChat2~\cite{li2023videochat} performs a progressive multi-modal training for three stages. In the first stage, it is trained to aling the visual encoder with a Querying Transformer (Q-Former)~\cite{blip2} which acts as an information bottleneck between the image and textual encoders and distill relevant information to the textual context. The second stage connects the visual encoder with a pretrained LLM. In the third stage, finetunes the model via instruction tuning, using 5 different tasks including: captioning, conversations,  visual question answering, reasoning and classification, with data coming from LLaVA~\cite{llava}, VideoChat~\cite{videochat}, VideoChatGPT~\cite{videochatgpt}, COCO Captions~\cite{coco}, WebVid~\cite{Bain2021FrozenIT}, YouCook~\cite{Das2013ATF}, OK-VQA~\cite{okvqa}, AOK-VQA~\cite{aokvqa}, DocVQA~\cite{docvqa}, CLEVR~\cite{clevr}, CLEVRER~\cite{clevrer} and NExT-QA~\cite{nextqa} among others.
We use this model under their MIT License.

\subsection{Datasets}
\label{app:datasets}

\paragraph{VideoCon.}
VideoCon is constructed by generating contrastive video captions and explanations for different subset of videos~\cite{Xu2016MSRVTTAL, wang2019vatex, hendricks-etal-2018-localizing}. 
This dataset contains seven misaligned types that include replacement of objects, actions, attributes, counts and relations, and adds hallucinations (i.e. unrelated but plausible information). We use this dataset under their MIT License.

\paragraph{VELOCITI.}
The duration of the video clips in the dataset is 10 seconds, and has dense text annotations on action and role descriptions.
The perception-based tests require discriminating video-caption pairs that share similar entities, and the binding tests require models to associate the correct entity to a given situation while ignoring the different yet plausible entities that also appear in the same video.
There are 1000 tests using 643 videos for Agent Iden, 1676 tests using 707 videos for Agent Bind, 418 tests using 270 videos for Agent Coref, 500 tests using 400 videos for Action Adv, 1625 tests using 590 videos for Action Bind, 500 tests using 411 videos for Action Mod, and 1908 tests using 669 videos for Chrono.  
We use this dataset under their Creative Commons Public Licenses.

\paragraph{SSv2-Temporal and SSv2-Events.}
SSv2-Temporal contains a list of 18 actions that require models to capture rich temporal information in the video, consisting of 216 (18 ×12) candidate videos for every text
action query.
SSv2-Events has 49 actions that consist two verbs in the action templates that are indicative of multiple events in the video,  consisting of 2888 (49×12) candidate videos for
every text action query.

%% file: tables/ssv2.tex

\begin{table}[!b]
\centering
\footnotesize
\begin{tabular}{@{}lcc@{}}
\toprule
Model        & SSv2-Temporal & SSv2-Events \\ \midrule
Random       & 7.3      & 3.3    \\
ImageBind \cite{girdhar2023imagebind}    & 10.5     & 5.5    \\
TACT \cite{bagad2023test}     & -  & 7.8 \\
mPLUG-V \cite{ye2023mplug} & 10.9 & 6.8 \\ 
VideoCon  \cite{bansal2024videocon}    & 15.2     & 11.4   \\ \midrule
Video-LLaVA  & 13.8     & 7.8    \\
+ Entail      & 17.5     & 7.5    \\
+ Entail+Mask &   18.0       &   9.9     \\
+ \method{}      & \textbf{19.5}     & 9.1    \\
+ \method{}+Mask  & 15.3     & \textbf{10.3}   \\ \midrule
\end{tabular}
\label{table:ssv}
\caption{Mean Average Precision (mAP) scores for the tested models in the zero-shot text-to-video retrieval tasks.}
\label{table:ssv2}
\end{table}

%% file: tables/different_mask_ratio.tex
\begin{table}[]
\centering
\footnotesize
\begin{tabular}{lccccccc>{\columncolor{gray!10}}c}
\toprule
\multirow{2}{*}{Model} & \multicolumn{3}{c}{\textbf{Agent Tests}}               & \multicolumn{3}{c}{\textbf{Action Tests}} & \textbf{Chrono} & \textbf{Avg} \\
& Iden          & Bind          & Coref         & Adv           & Bind          & Modif         &                         &                      \\
\midrule

\rowcolor[HTML]{F2F3F4} 
Video-LLaVA & 74.1 & 50.4 & \textbf{60.1} & 63.6 & 47.0 & 47.9 & 56.0 & 57.0  \\
\hdashline
\arrayrulecolor{black}
$+$ \method{} & 80.3 & \textbf{62.6} & 57.9 & \textbf{72.6} & \textbf{60.0} & \textbf{65.8} & 54.5 & 64.8 \\
$+$ \method{}+Mask 15\% & 77.9 & 58.6 & 58.4 & 71.4 & 57.7 & 61.7 & 57.4 & 63.3\\
$+$ \method{}+Mask 30\% & 81.4 & 60.3 & 54.3 & 70.0 & 58.0 & 65.4 & \textbf{59.5} & 64.1 \\
$+$ \method{}+Mask 45\% & \textbf{82.7} & 62.1 & 57.9 & 71.8 & 59.0 & 64.8 & 57.9 & \textbf{65.2} \\
$+$ \method{}+Mask 60\% & \textbf{82.7} & 62.1 & 56.5 & 70.6 & 57.8 & 62.3 & 55.7 & 64.0 \\

\bottomrule
\end{tabular}
\caption{Accuracy of Video-LLaVA jointly finetuned with \method{} and masking correction task using different masking ratio on VELOCITI (zero-shot).}
\label{table:mask_ratio}
\end{table}

%% file: tables/nle.tex
\begin{table}[]
\centering
\footnotesize
\begin{tabular}{lccccccc>{\columncolor{gray!10}}c}
\toprule
\multirow{2}{*}{Model} & \multicolumn{3}{c}{\textbf{Agent Tests}}               & \multicolumn{3}{c}{\textbf{Action Tests}} & \textbf{Chrono} & \textbf{Avg} \\
& Iden          & Bind          & Coref         & Adv           & Bind          & Modif         &                         &                      \\
\midrule

\rowcolor[HTML]{F2F3F4} 
Video-LLaVA & 74.1 & 50.4 & \textbf{60.1} & 63.6 & 47.0 & 47.9 & 56.0 & 57.0  \\
\hdashline
\arrayrulecolor{black}
$+$ Entail $+$ NLE & 77.1 & 60.0 & 58.6 & 66.8 & 55.9 & \textbf{66.3} & \textbf{57.9} & 63.2 \\
$+$ \method{} & 80.3 & \textbf{62.6} & 57.9 & \textbf{72.6} & \textbf{60.0} & 65.8 & 54.5 & 64.8 \\
$+$ \method{}+Mask & \textbf{82.7} & 62.1 & 57.9 & 71.8 & 59.0 & 64.8 & \textbf{57.9} & \textbf{65.2} \\

\bottomrule
\end{tabular}
\caption{Zero-shot accuracy on VELOCITI for models trained with the baseline entailment task, mixture of entailment and natural language explanation tasks, and our proposed \method~objective.}
\label{table:nle}
\end{table}

%% file: figures/qualitative_app.tex
\begin{figure*}[t]
\centering
\begin{subfigure}{0.48\textwidth}
\includegraphics[width=0.95\textwidth]
{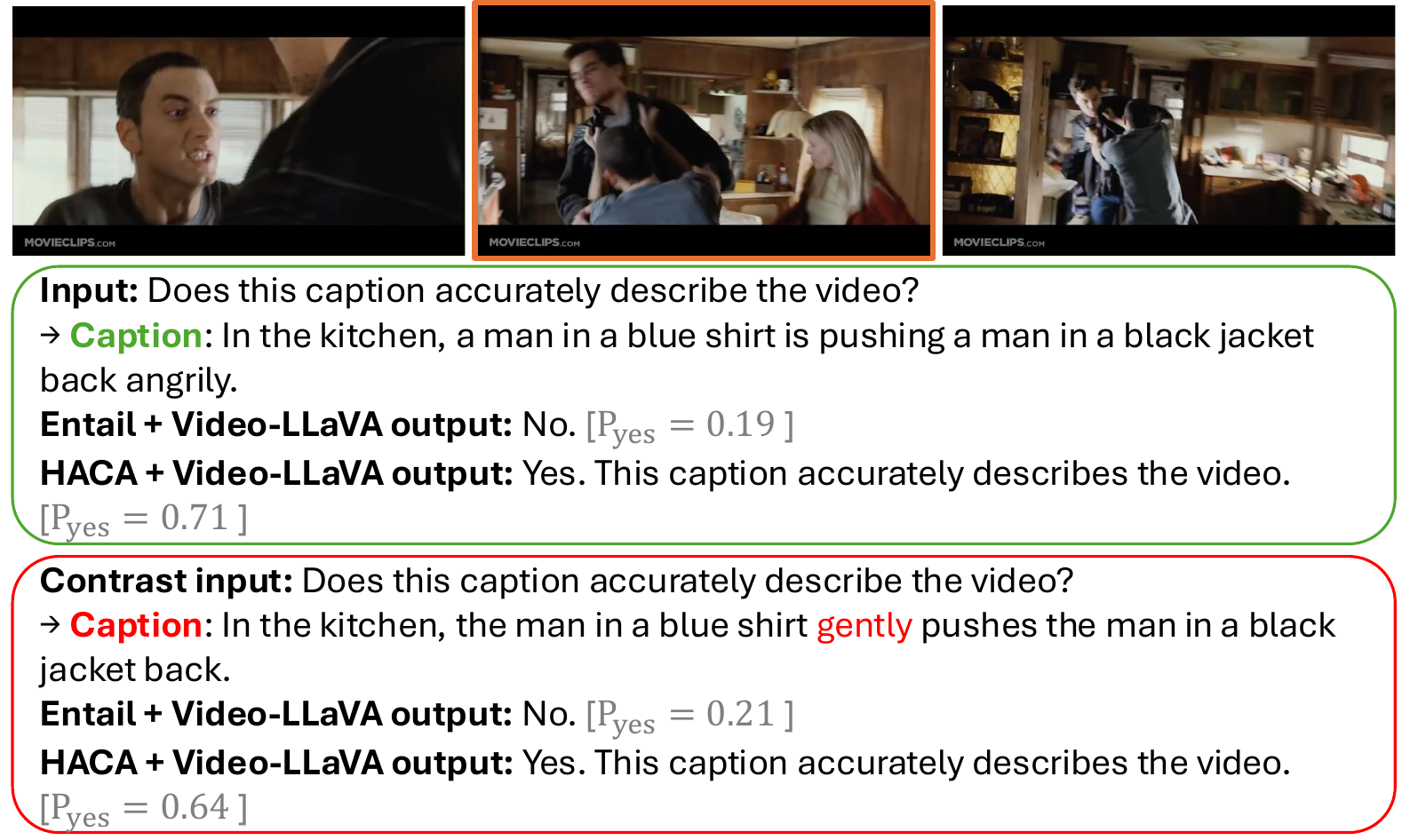}
\centering
         \caption{
         Success on binding, failure on correction: \method{} successfully assigns a higher entailment probability ($P_{yes}$) to the correct caption (top) compared to the incorrect one (bottom), outperforming the entailment-finetuned model in this regard. However, \method{} fails to produce a correction, as it erroneously indicates that the incorrect caption accurately describes the video.}
\end{subfigure}
\hfill
\begin{subfigure}{0.48\textwidth}
\includegraphics[width=0.95\textwidth]{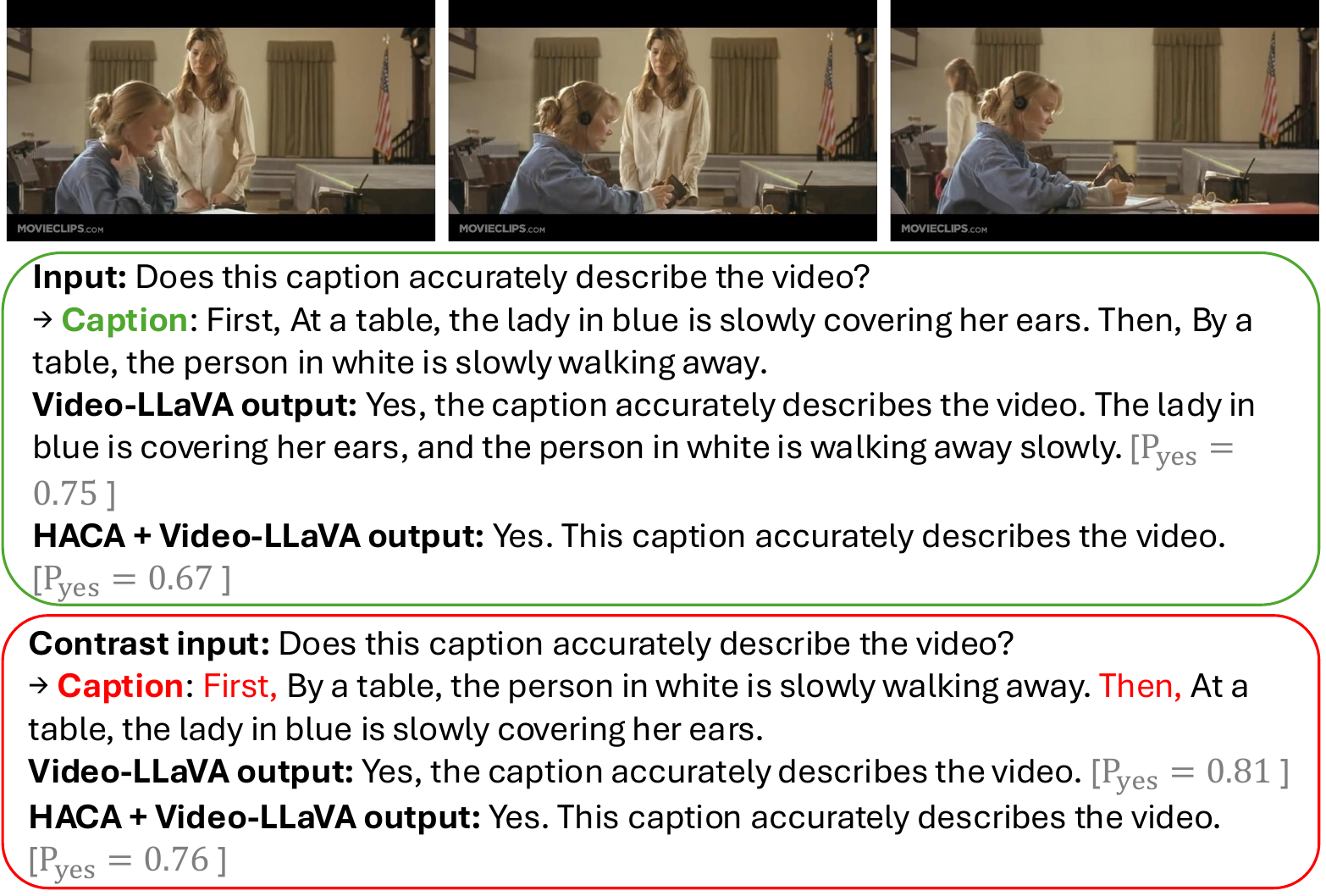}
\centering
         \caption{Failure on binding and correction: both \method{} and the pre-trained Video-LLaVA model incorrectly assign a higher entailment probability ($P_{yes}$) to the incorrect caption (bottom) than to the correct caption (top). Additionally, \method{} fails to provide a correction, mistakenly asserting that the incorrect caption accurately describes the video.}
\end{subfigure}
    \caption{Some successful and failure cases of \method{} and the other models on the VELOCITI dataset. The \textcolor{red}{red} color in text indicates the incorrect text description.}
    \label{fig:qualitative_app}
\vspace{-4mm}
\end{figure*}

%% file: sec/2_related_work2.tex
\section{Related Work}
\label{app:related_work}

\noindent\textbf{Alignment in Video-Language Models} is fundamental for the logical integration of video and textual information. To align both modalities, prior work has focused on pre-training models with different objectives to capture the temporal dynamics in video. 
While these self-supervised correction objectives are highly effective during pre-training~\cite{li2023lavender, wang2022ofa, zhu2024languagebind, ge2022bridging}, fine-tuning is typically required to adapt Video-LLMs to specific downstream tasks~\cite{videochat, videollama, bansal2024videocon} (e.g., classification, retrieval, or question answering).
However, these objectives rely on coarse-grained alignment labels and do not provide detailed feedback for resolving inconsistencies between video and language.

\noindent\textbf{Hallucination Correction} methods aim to mitigate the generation of content that does not align with the data a model was trained on, or the model describes content that does not exist in the provided input~\cite{huang2024opera}. Orthogonal to our proposed method, LURE~\cite{zhou2024analyzing} uses statistical analysis to identify and rectify errors in generated descriptions, addressing co-occurrence, uncertainty, and positional factors via masking. In our work, we randomly mask the video description so that the model is required to output the corrected sentence, which is also conditioned in the input video via visual entailment. 
\citet{yin2023woodpecker, wang2023paxion} uses external models and measures to correct hallucinations to be consistent with images or videos.
\citet{zhou-etal-2021-detecting, liu2023mitigating, xiao2024detecting, zhao-etal-2024-successfully} create a synthetic dataset to train a specialized model to detect and correct hallucinations.
\citet{dale2022detecting, huang2024opera} shows promising results on correcting hallucinations without an external model for machine translation and image captioning.
In our work, we investigate leveraging hallucination as a training objective to improve video-language alignment, by exploring the potential of using a video-LLM model itself to correct hallucinations through fine-tuning on a synthetic dataset.